\documentclass{article}

\usepackage{arxiv}

\usepackage[utf8]{inputenc} 
\usepackage[T1]{fontenc}    
\usepackage{hyperref}       
\usepackage{url}            
\usepackage{booktabs}       
\usepackage{amsfonts}       
\usepackage{nicefrac}       
\usepackage{microtype}      
\usepackage{lipsum}
\usepackage{graphicx}
\usepackage{amsmath}
\graphicspath{ {./images/} }
\usepackage{pgfplots}
\pgfplotsset{compat=1.18}
\usepackage{tikz}
\usepackage{hyperref}
\usepackage{algorithm}
\usepackage{algpseudocode}

\usepackage{fancyhdr} 
\pgfplotsset{
    akira_style/.style = {
        xtick = \empty,
        ytick pos=left,
        every y tick/.style={color=black, thin},
        tick align=outside,
        ylabel near ticks,
        xtick style={draw=none},
    }
}

\title{LADDER: Self-Improving LLMs Through Recursive Problem Decomposition}

\author{
Toby Simonds \\
 Tufa Labs\\
 \texttt{toby@tufalabs.ai} \\
  \And
Akira Yoshiyama \\
 Tufa Labs\\
 \texttt{akira@tufalabs.ai} \\
}
\begin{document}
\maketitle
\begin{abstract}

We introduce LADDER (Learning through Autonomous Difficulty-Driven Example Recursion), a framework which enables Large Language Models to autonomously improve their problem-solving capabilities through self-guided learning by recursively generating and solving progressively simpler variants of complex problems. Unlike prior approaches that require curated datasets or human feedback, LADDER leverages a model's own capabilities to generate easier question variants. We demonstrate LADDER's effectiveness on the subject of mathematical integration, improving Llama 3.2 3B's accuracy from 1\% to 82\% on undergraduate-level problems and enabling Qwen2.5 7B Deepseek-R1 Distilled to achieve 73\% on the MIT Integration Bee qualifying examination. We also introduce TTRL (Test-Time Reinforcement Learning), where we perform reinforcement learning on variants of test problems at inference time. TTRL enables Qwen2.5 7B Deepseek-R1 Distilled to achieve a state-of-the-art score of 90\% on the MIT Integration Bee qualifying examination, surpassing OpenAI o1's performance. These results show how self-directed strategic learning can achieve significant capability improvements without relying on architectural scaling or human supervision.

\end{abstract}

\section{Introduction}

Reinforcement Learning (RL) has emerged as a highly effective approach for training Large Language Models (LLMs), yet its success hinges critically on the availability of appropriate training tasks \cite{lambert2025tulu3pushingfrontiers, openaio1, ouyang2022traininglanguagemodelsfollow, wang2023selfconsistencyimproveschainthought}. A fundamental challenge lies in obtaining verifiable tasks that match the model's current capabilities. For RL to be effective, tasks must form a gradient of difficulties that allows for incremental learning progress \cite{narvekar2020curriculumlearningreinforcementlearning}. When tasks exceed the model's current abilities, the training process not only stalls but can lead to catastrophic collapse, resulting in degraded performance. This challenge is particularly acute in domains requiring complex reasoning, where the gap between simple and advanced tasks can be substantial \cite{korbak2022reinforcementlearningdistributionmatching, wang2023selfconsistencyimproveschainthought}.

We propose Learning through Autonomous Difficulty-Driven Example Recursion (LADDER), a framework that enables LLMs to autonomously improve their problem-solving capabilities through strategic self-guided learning. The key insight is that models can bootstrap their own learning by recursively generating and solving progressively simpler variants of complex problems. For each challenging problem, LADDER prompts the model to create multiple easier variants, forming a natural difficulty gradient. This process continues recursively, with each variant spawning simpler sub-variants, until reaching problems the model can reliably solve. The solutions to these simpler problems then provide stepping stones for tackling progressively harder variants. We demonstrate that this self-bootstrapping approach achieves dramatic improvements beyond what's possible through standard techniques like pass@k sampling - enabling models to reliably solve problems that were previously far beyond their capabilities.



Unlike previous approaches requiring carefully curated datasets or human feedback, LADDER leverages the model's existing capabilities to create a natural difficulty gradient, allowing for systematic improvement through reinforcement learning with verifiable rewards. The framework requires only a reliable verification mechanism - in our case, numerical integration for checking solutions. This enables the model to assess its own progress and guide its learning trajectory without human intervention.

We demonstrate LADDER's effectiveness on mathematical integration tasks, achieving remarkable improvements across multiple benchmarks. Using this approach, we improve a Llama 3B model's accuracy from 1\% to 82\% on undergraduate-level integration problems. When applied to the challenging 2025 MIT Integration Bee examination, LADDER enables a 7B parameter model to achieve 73\% accuracy, significantly outperforming much larger models, such as GPT-4o (42\%), and typical human performance (15-30\%). These results showcase how strategic problem decomposition and verified self-learning can achieve substantial capability improvements without relying on architectural scaling or human supervision.

Building on LADDER's self-improvement framework, we propose Test-Time Reinforcement Learning (TTRL), a novel approach that extends these principles to inference time. TTRL dynamically generates problem variants during test-time and applies reinforcement learning to refine the model's solutions, effectively creating a micro-learning process for each test instance. By leveraging the same verification mechanisms used in training, TTRL enables the model to further improve its performance. When applied to the 2025 MIT Integration Bee, TTRL boosts accuracy from 73\% - with just LADDER - to 90\%, demonstrating how scaling test-time compute through strategic problem decomposition can yield substantial performance improvements. We achieve state of the art accuracy, outperforming significantly larger models, such as OpenAI's o1.

Thus, we make the following contributions:
\begin{itemize}
    \item We propose a novel framework for autonomous model improvement through recursive problem decomposition and self-guided learning via reinforcement learning with GRPO.
    \item We develop a systematic method for generating and verifying problem variants that create natural difficulty gradients, requiring only numerical verification.
    \item We demonstrate significant empirical improvements on mathematical reasoning tasks, improving a Llama 3B model from 2\% to 82\% on undergraduate integration problems and achieving 73\% accuracy on the MIT Integration Bee with a 7B model, matching SoTA performance
    \item We introduce Test-Time Reinforcement Learning (TTRL), a method for scaling compute at inference time through variant generation and reinforcement learning, boosting performance on the MIT Integration Bee from 73\% to 90\%.
\end{itemize}

\section{Related Work}
\label{sec:related}

\paragraph{Self-Improvement and Automated Curriculum Generation in LLMs.}
Haluptzok et al. (2023) present a self-play setup where a code-focused LLM continually invents and solves new programming puzzles, checks solutions with a Python interpreter, and fine-tunes on correct results \cite{haluptzok2023languagemodelsteachprogram}. We build upon the generate → solve → verify → learn cycle, adding a more explicit curricular element to guide the model’s step-by-step improvement.

Recent advances such as STaR (Self-Taught Reasoner) have illustrated that LLMs can improve their reasoning skills by learning from their own generated chain-of-thoughts—effectively acting as both student and teacher \cite{zelikman2022starbootstrappingreasoningreasoning}. Unlike traditional self-play techniques that require competitive dynamics, these methods harness the model’s intrinsic capacity to generate and evaluate intermediate reasoning steps, automatically curating training examples that progressively reduce task complexity. Our work extends this line of research by systematically transforming unsolvable problems into tractable sub-problems, thereby constructing an end-to-end self-improving framework that requires no human annotation or intervention.

\paragraph{Test-Time Compute Scaling and Reasoning Improvements.}
Recent work has explored various approaches to improving model performance through increased compute at inference time. "Let's Verify Step by Step" \cite{lightman2023letsverifystepstep} demonstrated that breaking down verification into explicit steps and allocating additional computation to each verification component can significantly improve performance. Similarly, Tree of Thoughts \cite{yao2023treethoughtsdeliberateproblem} showed how systematically exploring multiple reasoning paths during inference can lead to better problem-solving capabilities. These approaches typically focus on increasing the length and deliberation of model outputs, either through structured prompting, such as s1, or systematic exploration of solution spaces, such as self-consistency \cite{muennighoff2025s1simpletesttimescaling, wang2023selfconsistencyimproveschainthought}. Our work differs fundamentally from these approaches by introducing dynamic learning at test-time - rather than just increasing reasoning length or exploring multiple paths, TTRL enables the model to actually improve its capabilities through practice on related problems. This represents a shift from static inference-time techniques to dynamic adaptation through targeted learning.

\paragraph{Reinforcement Learning Approaches for LLM Self-Improvement.}
In recent years, reinforcement learning has emerged as a pivotal strategy for enabling language models to self-improve through iterative self-correction and feedback. Notably, OpenAI’s o1 model uses reinforcement learning to refine its chain‐of‐thought reasoning—learning to “think” step by step \cite{openaio1}. Similarly, DeepSeek’s R1 model leverages a reinforcement learning framework that minimizes human supervision by generating and self-evaluating its own training data \cite{deepseekai2025deepseekr1incentivizingreasoningcapability}. 

Importantly, recent work has shown that RL approaches may generalize much better than Supervised Fine-Tuning (SFT), which appears to memorize instead \cite{chu2025sftmemorizesrlgeneralizes, xie2025logicrlunleashingllmreasoning}. RL approaches have also shown to be highly effective in generalizing to out-of-distribution tasks, particularly in competitive mathematics \cite{xie2025logicrlunleashingllmreasoning}. These results strongly suggest that reinforcement learning offers a more promising path to self-improvement than supervised fine-tuning.

These innovations underscore a paradigm shift—from externally curated curricula and human-labeled data toward autonomous, feedback-driven self-improvement in language models.

\section{Methodology}

\begin{figure}[t]
\centering
\includegraphics[width=0.8\textwidth]{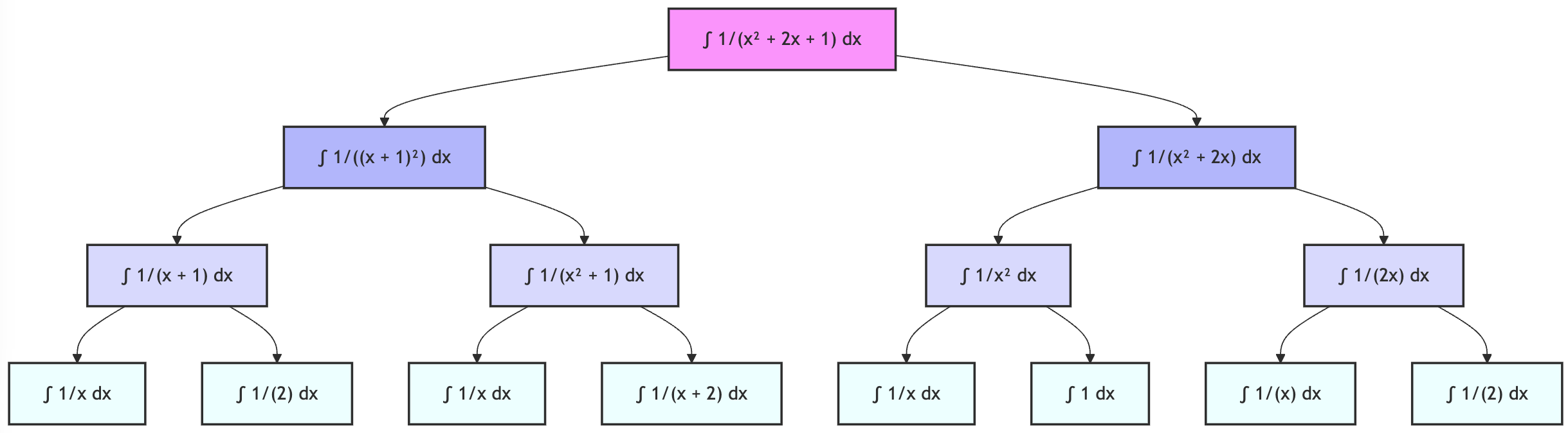}
\caption{Example of variant generation for an integration problem. Each level represents progressively simpler variants of the original problem. The tree structure ensures each variant has exactly one parent, maintaining a clear progression of difficulty.}
\label{fig:integral-variants}
\end{figure}

\subsection{Algorithm Design (LADDER \& TTRL)}

\subsubsection{LADDER}\label{sec:ladder}

LADDER is a structured framework designed to enhance LLM problem-solving capabilities through recursive problem decomposition and reinforcement learning. We demonstrate the effectiveness of LADDER in the domain of mathematical integration. LADDER consists of the following components:

\begin{itemize}
    \item \textbf{Variant Generation}: A structured method for generating trees of progressively simpler variants of complex problems, establishing a natural difficulty gradient.
    \item \textbf{Solution Verification}: A numerical integration method to verify the solution of an integral.
    \item \textbf{Reinforcement Learning}: The protocol used to train the base model on the variant trees.
\end{itemize}

LADDER is a two-step algorithm. First, the training set of integrals is collected and a tree of variants is generated for each integral as described in section \ref{sec:variants}. Second, we perform the reinforcement learning protocol described in section \ref{sec:rl} on a base model using the set of variant trees as the training set. The resulting model should then have significantly improved and generalized mathematical integration capabilities, and can then be prompted to solve new integrals. See Algorithm \ref{algorithm1} for LADDER pseudocode. \(V_{\text{LADDER}}\) is the set of variants generated for all questions in the train set of integrals and \(a_i\) is the answer to the \(i\)th test question.

The number of variants \(N\) generated per question is a hyperparameter, which we vary for different experiments. We experiment with LADDER in sections \ref{sec:Llama} and \ref{sec:mit}.

\subsubsection{Variant Generation}\label{sec:variants}
The variant generation process required careful design to ensure sufficient diversity and appropriate difficulty scaling. We found that naive approaches to variant generation - simply asking the model to generate easier versions - led to repetitive patterns and limited diversity. Through experimentation, we developed a multi-stage process that dramatically improved the quality and variety of generated variants.

\begin{minipage}{0.48\textwidth}
\begin{algorithm}[H]
    \centering
    \caption{LADDER}\label{algorithm1}
    \begin{algorithmic}[1]
        \State $Q_{\text{train}}, Q_{\text{test}} \gets \text{train and test sets of integrals}$
        \State $N \gets \text{number of variants to generate}$
        \State $\text{G}(q,j) : \text{generates the \(j\)th variant of integral \(q\)}$
        \State \scalebox{0.85}{$V_{\text{LADDER}} \gets \{v_{i,j} \in \text{G}(Q_{\text{train}}[i], j) \;|\; 1\le i\le |Q_\text{train}|,1\le j\le N\}$}
        \State $\pi_{\theta_{\text{LADDER}}} \gets \text{GRPO}(\pi_{\theta_{\text{base}}}, V_{\text{LADDER}})$
        \For{$i\gets 1, |Q_{\text{test}}|$}
        \State $a_{i} \sim \pi_{\theta_{\text{LADDER}}}(\cdot \;|\;Q_{\text{test}}[i])$ 
        \EndFor
    \end{algorithmic}
\end{algorithm}
\end{minipage}
\hfill
\begin{minipage}{0.48\textwidth}
\begin{algorithm}[H]
    \centering
    \caption{TTRL}\label{algorithm2}
    \begin{algorithmic}[1]
        \State $Q_{\text{test}} \gets \text{test set of integrals}$
        \State $N \gets \text{number of variants to generate}$
        \State $\text{G}(q,j) : \text{generates the \(j\)th variant of integral \(q\)}$
        \For{$i\gets 1, |Q_{\text{test}}|$}
        \State $ V_{\text{TTRL},i} \gets \{v_{j} \in \text{G}(Q_{\text{test}}[i], j) \:|\:1\le j \le N\}$
        \State $\pi_{\theta_{\text{TTRL}},i} \gets \text{GRPO}(\pi_{\theta_{\text{base}}}, V_{\text{TTRL},i})$
        \State $a_{i} \sim \pi_{\theta_{\text{TTRL}},i}(\cdot \;|\;Q_{\text{test}}[i])$ 
        \EndFor
    \end{algorithmic}
\end{algorithm}
\end{minipage}

\vspace{\baselineskip}
\vspace{\baselineskip}

First, we used the base model to generate an extensive set of mathematical transformations that could be applied to integrals, categorized by their impact on problem difficulty. These included operations ranging from simple (e.g., reducing exponent values, simplifying denominators) to more complex (e.g., introducing nested functions, combining multiple function types). This transformation library served as a foundation for guiding variant generation.

Second, for each integral, we randomly sampled 3-5 transformations and provided them as explicit suggestions to the variant generation model. Critically, we found that generating variants in batches of 10 per prompt significantly improved diversity - smaller batch sizes led to more repetitive outputs, while larger batches reduced quality. To further enhance variation, we employed two key techniques:

\begin{enumerate}
    \item \textbf{Temperature cycling:} We dynamically varied the sampling temperature between 0.8 and 1.4 across prompts. This helped balance between creativity and mathematical validity.
    \item \textbf{Persona-based prompting:} We prompted the model to adopt different mathematical perspectives (e.g., "think like Euler focusing on series", "approach like Gauss looking for patterns").
\end{enumerate}

The combination of batch generation, varied temperatures, and rotating personas proved crucial led to significantly more diverse variants. Without them the model would often converge to generating very similar  and often repeating variants.


Third, to ensure sufficient coverage of the difficulty space, we applied this generation process recursively, generating a tree of variants, each a simpler integral than its parent. We found that this also helped improve variant diversity. This recursive approach helped build natural difficulty gradients - each problem could spawn multiple easier variants, which in turn could generate even simpler problems. We capped tree depth at three to maintain problem relevance. The following is an example of one root-to-leaf path in a variant tree:

\begin{align*}
\text{Original problem:}\quad
& \int \frac{x^2 + 1}{x^4 + 2x^2 + 1} dx \\[2pt]
\text{Level 1 variant:}\quad
& \int \frac{x^2}{x^4 + 1} dx \\[2pt]
\text{Level 2 variant:}\quad
& \int \frac{1}{x^2 + 1} dx \\[2pt]
\text{Level 3 variant:}\quad
& \int \frac{1}{x^2} dx \\[2pt]
\end{align*}

Quality control presented a significant challenge in our variant generation process. While we prompted the model to generate "easier" or "equivalent" variants, the actual difficulty of the resulting integrals often deviated substantially from the intended level. Small perturbations in coefficients or function composition could transform seemingly simple integrals into much harder ones - for instance, changing a coefficient from 1 to 2 in a rational function could introduce complex roots that make the integral significantly more challenging. 

For our Llama experiments, approximately 8\% of generated variants were effectively unsolvable (i.e. it took a symbolic system more than 5 seconds to solve), and many more were substantially harder than their parent problems despite being intended as simpler variants. While these harder variants were naturally ignored during training through zero reward signals, they consumed significant computational resources in verification attempts and represented wasted generation capacity. Future work could explore methods to better constrain variant generation to maintain intended difficulty levels while preserving mathematical diversity.

We apply the above three-step variant generation method to generate the sets of variant integrals in both LADDER and TTRL. \(V_{\text{LADDER}}\) is the set of variants for all questions in the train set \(Q_\text{train}\) and \(V_{\text{TTRL},i}\) is the set of variants for the \(i\)th question in the test set, \(Q_\text{test}[i]\). See Algorithms \ref{algorithm1} and \ref{algorithm2}.

\subsubsection{Solution Verification}\label{sec:solverif}
Integral variant generation occurs at the beginning of LADDER. Throughout the process of performing RL in LADDER we must also perform solution verification. In order to do so, we developed a robust numerical verification framework that balanced accuracy with computational efficiency. The key challenge was ensuring reliable verification across different types of integrals while handling edge cases and potential numerical instabilities.

For each candidate solution to an integral, we employed a multi-point numerical comparison approach. We randomly sampled five points from the domain [-10, 10] and evaluated both the candidate solution and the original integral over small intervals of length 0.1 centered at these points. This interval length was chosen empirically to minimize numerical integration errors while maintaining sensitivity to local solution behavior. The use of multiple small intervals, rather than a single large one, helped detect both local and global errors in the solutions.

Our verification protocol included several key components:
\begin{itemize}
   \item \textbf{Singularity handling}: When a sampled point was near a singularity (detected through rapid value changes or numerical overflow), we automatically resampled to a new point. This adaptive sampling ensured robust verification even for functions with challenging behavior.
   
   \item \textbf{Numerical precision}: Solutions were deemed correct if their values exhibited a relative difference of $10^{-2}$ or less compared to numerical integration results across all test intervals. We found empirically that this tolerance level effectively balanced false positives and negatives.

   \item \textbf{Timeout management}: To handle computationally intensive integrals, we implemented a 2-second timeout for each verification attempt. If timeout occurred, new evaluation points were sampled, with up to three retries before marking the solution as unverifiable.
   
   \item \textbf{Edge case detection}: Early iterations revealed that models could exploit verification weaknesses by outputting trivial solutions (like the integration symbol itself) or degenerate forms. We added specific filters to detect and reject such cases, ensuring solutions demonstrated genuine understanding.
\end{itemize}

All numerical integrations were performed using adaptive quadrature methods, with automatic precision adjustment based on the integrand's complexity. This approach provided reliable results even for oscillatory and highly nonlinear functions, while maintaining reasonable computational efficiency.

We note, however, that the numerical verifier may not be 100\% accurate. We opted for a numerical approach rather than a symbolic one because many problems beyond integration can be numerically verified despite lacking a symbolic solver.

We apply the above solution verification in our RL Protocol in section \ref{sec:rl} and experiments in sections \ref{sec:Llama}, \ref{sec:mit} and \ref{sec:ttrltest}. In our MIT experiments, final benchmark results are verified directly against the official solutions.

\subsubsection{Reinforcement Learning Protocol}\label{sec:rl}

We decided to employ Group Relative Policy Optimization (GRPO) in both LADDER and TTRL \cite{shao2024deepseekmathpushinglimitsmathematical}. GRPO does not use a separate critic model and instead estimates the baseline from group scores, improving efficiency and reducing memory. For each question \(q\), GRPO samples a group of outputs \({o_1, o_2,...,o_G}\) from the old policy \(\pi_{\theta_{old}}\)  and then optimizes the policy model \(\pi_\theta\) by maximizing the following objective:

\begin{equation}
\begin{aligned}[t]
&\hspace{-2em}
J_{\mathrm{GRPO}}(\theta)
= \mathbb{E}\Bigl[
  q \sim P(Q),\;
  \{o_i\}_{i=1}^G \sim \pi_{\theta_{\mathrm{old}}}(O \mid q)
\Bigr]
\\[6pt]
&\frac{1}{G}
\sum_{i=1}^G
\Bigl(
  \min\Bigl(
    \frac{\pi_\theta(o_i \mid q)}{\pi_{\theta_{\mathrm{old}}}(o_i \mid q)}\,A_i,\;
    \mathrm{clip}\Bigl(
      \frac{\pi_\theta(o_i \mid q)}{\pi_{\theta_{\mathrm{old}}}(o_i \mid q)},
      \,1 - \varepsilon,\,
      \,1 + \varepsilon
    \Bigr)\,A_i
  \Bigr)
  \;-\;
  \beta\,D_{\mathrm{KL}}\bigl(\pi_\theta \,\|\, \pi_{\mathrm{ref}}\bigr)
\Bigr)
\end{aligned}
\end{equation}

\begin{equation}
D_{\mathrm{KL}}(\pi_\theta \,\|\, \pi_{\mathrm{ref}})
=
\frac{\pi_{\mathrm{ref}}(o_i \mid q)}{\pi_\theta(o_i \mid q)}
\;-\;
\log
\Bigl(
\frac{\pi_{\mathrm{ref}}(o_i \mid q)}{\pi_\theta(o_i \mid q)}
\Bigr)
\;-\;
1
\end{equation}

where \( \epsilon \) and \( \beta \) are hyperparameters, and \( A_i \) is the advantage, computed using a group of rewards \({r_1, r_2,..., r_G}\) corresponding to the outputs within each group:

\begin{equation}
A_i
=
\frac{
r_i
-
\mathrm{mean}\bigl(\{r_1, r_2, \dots, r_G\}\bigr)
}{
\mathrm{std}\bigl(\{r_1, r_2, \dots, r_G\}\bigr)
}
\end{equation}


We adopted a simple, rule-based reward model. The reward model is kept simple in order to be straightforward to apply to other verifiable domains in the future. The reward model consists of two rewards:

\begin{itemize}
    \item Accuracy reward: Using the solution verification method described in \ref{sec:solverif}, we evaluate whether the response is correct or not. The model is required to provide the final answer in a specified format (i.e. in <ANSWER> </ANSWER> tags), enabling reliable rule-based verification of correctness.
    \item Format reward: In addition to the accuracy reward model, we employ a format reward that encourages the model to put its answer in <ANSWER> </ANSWER> tags.
\end{itemize}

We perform the RL runs with a KL divergence coefficient of 0.001. Training batch size and number of epochs differ among our experiments.

\subsubsection{Test-Time Reinforcement Learning}\label{sec:ttrl}

Test-Time Reinforcement Learning (TTRL) extends our LADDER framework to inference time, enabling dynamic adaptation to challenging problems directly during testing. Upon encountering a difficult integration problem, TTRL generates a focused set of related variants and conducts targeted reinforcement learning specifically for that problem. By recursively decomposing the challenging problem into simpler variants and learning from this focused curriculum, TTRL allows the model to rapidly develop problem-specific expertise without hand-crafting variants or architectural modifications. This provides a novel approach to scaling compute at test-time - rather than simply increasing output sampling or model size, TTRL leverages compute to dynamically improve the model's problem-solving capabilities through focused practice on variants of the specific challenge at hand.

The same three components of LADDER are also used in Test-Time Reinforcement Learning (TTRL). For each question at test-time, there are two steps. First, we generate a tree of variants for the test question at hand. Second, we perform the reinforcement learning protocol described in section \ref{sec:rl} on a base model using the variant tree as the training set. The resulting model should then have significantly improved mathematical integration capabilities tuned to the test question at hand. The test question is then answered using the tuned model, and finally the model is rolled back to its original parameters for the next test question. See Algorithm \ref{algorithm2} for TTRL pseudocode, where \(V_{\text{TTRL},i}\) is the set of variants generated for the \(i\)th test question and \(a_i\) is the answer to the \(i\)th test question.

As in LADDER, the number of variants generated per question is a hyperparameter, which we vary for different experiments. We experiment with TTRL in section \ref{sec:ttrltest}.

\subsection{Experiments}

\subsubsection{Llama 3B Experiments}\label{sec:Llama}

We first experiment with LADDER using a Llama 3.2 3B parameter model as our base architecture to allow us to more quickly run ablation experiments. To establish our evaluation dataset, we developed a comprehensive collection of 110 indefinite integration problems, combining questions sourced from university-level mathematics curricula with synthetically generated problems using GPT-4o. To ensure appropriate difficulty calibration, we benchmarked each problem to verify it was solvable by GPT-4o but beyond the capabilities of the base Llama 3.2 3B model. This was done purposefully to benchmark performance on problems that can still be solved by an LLM but are outside the scope of Llama 3.2 3B.

\begin{figure}[h]
\centering
\begin{minipage}{0.48\textwidth}
\begin{tikzpicture}
\begin{axis}[
    xtick pos=left,
    ytick pos=left,
    every x tick/.style={color=black, thin},
    every y tick/.style={color=black, thin},
    tick align=outside,
    xlabel near ticks,
    ylabel near ticks,
    xtick style={draw=none},
    axis on top,     
   width=7cm,
   height=7cm,
   grid=major,
   xlabel={Steps},
   ylabel={Test Score (\%)},
   xmin=0,
   xmax=350,
   ymin=0,
   ymax=100,
   xtick distance=100,
   ytick distance=20,
   legend pos=south east,
   legend style={draw=none}
]
\addplot[
   blue,
   thick,
] coordinates {
    (0,1)
    (10,4)
    (20,23)
    (30,27)
    (40,18)
    (50,32)
    (60,34)
    (70,37)
    (80,37)
    (90,42)
    (100,42)
    (110,41)
    (120,51)
    (130,50)
    (140,56)
    (150,59)
    (160,65)
    (170,59)
    (180,70)
    (190,76)
    (200,73)
    (210,73)
    (220,76)
    (230,77)
    (240,85)
    (250,86)
    (260,83)
    (270,82)
    (280,81)
    (290,80)
    (300,82)
    (310,84)
    (320,86)
    (330,84)
    (340,81)
    (350,84)
};
\end{axis}
\end{tikzpicture}
\caption{LADDER training progression.}
\label{fig:training-progression}
\end{minipage}
\hfill
\begin{minipage}{0.48\textwidth}
\begin{tikzpicture}
\begin{axis}[
    xtick pos=left,
    ytick pos=left,
    every x tick/.style={color=black, thin},
    every y tick/.style={color=black, thin},
    tick align=outside,
    xlabel near ticks,
    ylabel near ticks,
    xtick style={draw=none},
    axis on top,   
    width=7cm,
    height=7cm,
    grid=major,
    xlabel={Number of Generated Variants},
    ylabel={Test Score (\%)},
    xmin=0,
    xmax=10,
    ymin=0,
    ymax=100,
    xtick={0,3,6,9},
    xticklabels={0,3k,6k,9k},
    ytick={0,20,40,60,80,100},
    legend pos=south east,
    legend style={draw=none}
]
\addplot[
    blue,
    thick,
    mark=*,
] coordinates {
    (0,0)
    (3,52)
    (6,72)
    (9,82)
};
\end{axis}
\end{tikzpicture}
\caption{Performance scaling w/ variant generation.}
\label{fig:scaling-performance}
\end{minipage}
\end{figure}

We randomly split this collection into a training set of 10 problems and a test set of 100 problems. we generated approximately 500 variants for each of our 10 training problems. To validate the diversity of our variant set, we performed exact matching against our test set and found minimal overlap (only 5 matches among 5,000 variants). The small training set was deliberately chosen to demonstrate our method's effectiveness in generating useful variants from limited examples. Approximately half of the problems were sourced from curriculum textbooks, with the remainder synthetically generated by GPT-4o. Notably, the synthetic problems were indistinguishable from curriculum-sourced ones.

During training, models were prompted to express solutions in sympy algebraic notation without integration constants, ensuring consistent evaluation across different but mathematically equivalent expressions of the same solution. This standardization was crucial for reliable verification of solution correctness during the training process.

We performed two RL experiments with Llama 3.2 3B. The first experiment trained on only the 10 question training set (RL w/o variants). The second experiment trained on only variants generated from the 10 question training set (RL w/ variants, i.e. LADDER). Both runs used identical hyper parameters and were continued until performance plateaued.

\subsubsection{MIT Integration Bee (LADDER)}\label{sec:mit}
Building off the findings from our Llama 3B experiemtns, we extended our methodology to the 2025 MIT Integration Bee qualifying examination, an annual competition that attracts both undergraduate and graduate students from MIT, with participants often having competitive mathematics backgrounds. The qualifying exam, which serves to select 16 finalists, contains 20 questions and has a typical qualifying threshold of 73\%, though most participants score between 15-30\%, with 50\% to 73\% out of 20 considered strong performance. Students are given 20 minutes to attempt the entire exam.

Using the DeepSeek-R1 distilled Qwen 2.5 7B model, we applied our variant generation approach to historical qualifying exams from 2010-2024. Our variant generation followed a two-level tree structure - first generating variants from each source problem, then generating additional variants from those first-level variants. We capped recursion at depth two, as preliminary experiments showed third-level variants became trivially simple and lost mathematical relevance to the original problems. With this approach we generated 9,000 variants.

Through experimentation with different difficulty distributions, we settled on prompting the model to generate 70\% easier and 30\% equivalent variants than their parent problems. While we initially attempted to include more challenging variants, we found this skew toward easier problems created more effective learning trajectories. Notably, many "equivalent" variants still introduced novel mathematical patterns while maintaining similar difficulty levels, contributing to the model's generalization capabilities.

We verified that none of the 2025 exam questions appeared in our variant set, though we note this precaution was primarily for methodological rigor rather than necessity, as our model never accessed the test set or solutions during training. We apply the same hyperparameters as in the 3B experiments, except for modifying the batch size to 128.

\subsubsection{MIT Integration Bee (LADDER + TTRL)}\label{sec:ttrltest}

After applying LADDER, there remain certain questions which the tuned model continues to fail to answer. For each of these questions we further apply TTRL. For each unsolved problem, we generate a tree of variants following the same process illustrated in Figure \ref{fig:integral-variants}, but limited to two levels of depth and approximately 800 total variants. Using these problem-specific variants, we conduct 100 steps of reinforcement learning with identical RL parameters as in the MIT Integration Bee LADDER setup. The problem is considered "solved" if the model produces a correct solution at any point during this process, as verified by our numerical integration framework.

TTRL’s approach of on-the-fly data synthesis means it can handle novel integrals better by expanding its training set - focused on that problem - at test time. This is especially important for challenges such as the MIT Integration Bee, where many integrals are cleverly constructed and might be out-of-distribution relative to standard calculus problems. Further, TTRL provides a systematic approach to scaling performance through additional compute. Rather than simply increasing sampling or temperature parameters, TTRL enables active improvement of problem-solving capabilities through focused practice at test-time. 

\section{Results}




\begin{figure}[h]
\centering
\begin{tikzpicture}
\begin{axis}[
    akira_style,
    ybar,
    bar width=20pt,
    height=6cm,
    symbolic x coords={Base Model (pass@1),Base Model (pass@10),RL w/o variants,LADDER (RL w/ variants)},
    xtick=data,
    xticklabel style={rotate=45,anchor=east},
    ylabel={Test Score (\%)},
    nodes near coords,
    nodes near coords align={above},
    every node near coord/.append style={font=\small,text=black},
    ymin=0,
    ymax=100,
    ]
\addplot coordinates {
    
    (Base Model (pass@1),1)
    (Base Model (pass@10),2)
    (RL w/o variants,3)
    (LADDER (RL w/ variants),82)
};
\end{axis}
\end{tikzpicture}
\caption{Comparison of LLM scores by training approach.}
\label{fig:llmcomparison}
\end{figure}
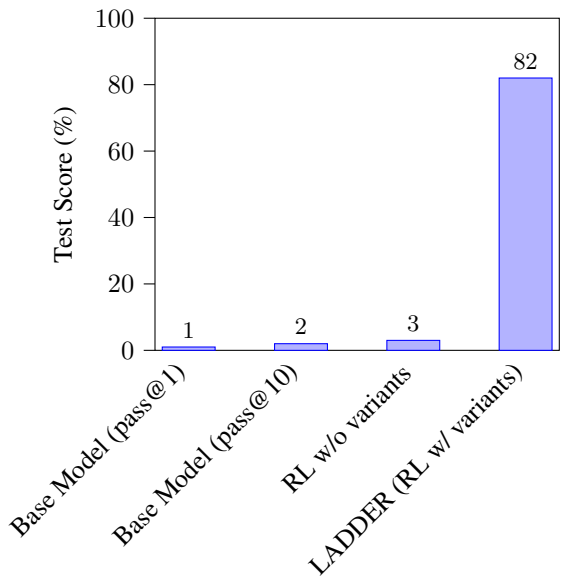

\subsection{Llama 3B Experiments}

Our experiments evaluated whether generating easier variants could enable effective reinforcement learning in mathematical integration tasks on Llama 3.2 3B. We tested four conditions: base model with pass@1 samping, base model with pass@10 sampling, RL without variants (only on the training set), and LADDER (RL with variants from the training set).

In Figure \ref{fig:training-progression}, we observe a consistent and rapid improvement from step 0 to 250, with the model's performance quickly ascending before ultimately plateauing. This swift progression suggests an effective learning mechanism, characterized by substantial performance gains that stabilize around 82\% accuracy as the model approaches its performance ceiling.

LADDER achieved 82\% accuracy on the test set, significantly outperforming both the base model with pass@1 sampling (1\%) and with pass@10 sampling (2\%). RL without variants consistently failed, with performance never exceeding 3\% before collapsing to 0\% within 30 training steps, highlighting the necessity of a smooth difficulty gradient for successful training (see Figure \ref{fig:llmcomparison}).


Analysis of the remaining unsolved problems reveals nuanced challenges in the model's mathematical integration capabilities. The model demonstrates difficulty with integrals requiring multiple technique combinations and novel substitution patterns. These complex scenarios highlight limitations in the current approach, suggesting that while the variant generation successfully teaches individual techniques, synthesizing these methods for more intricate problems remains a challenge.


\begin{figure}[h]
\centering
\begin{tikzpicture}
\begin{axis}[
    akira_style,
    ybar,
    bar width=15pt,
    width=\textwidth,
    height=8cm,
    symbolic x coords={ Claude-3.5,  GPT-4o, Base Model, LADDER,  o1},
    xtick=data,
    xticklabel style={rotate=45,anchor=east},
    ylabel={Score (\%)},
    ymin=30,
    ymax=85,
    nodes near coords,
    nodes near coords align={above},
    every node near coord/.append style={font=\small,text=black},
    ]
\addplot coordinates {
    
    (Claude-3.5,32)
    
    (GPT-4o,42)
    (Base Model,50)
    (LADDER,73)
    (o1,80)

};
\end{axis}
\end{tikzpicture}
\caption{Comparison of LLM scores on the 2025 MIT Integration Bee (7B). Average across 3 runs.} 
\label{fig:mit-perf}
\end{figure}
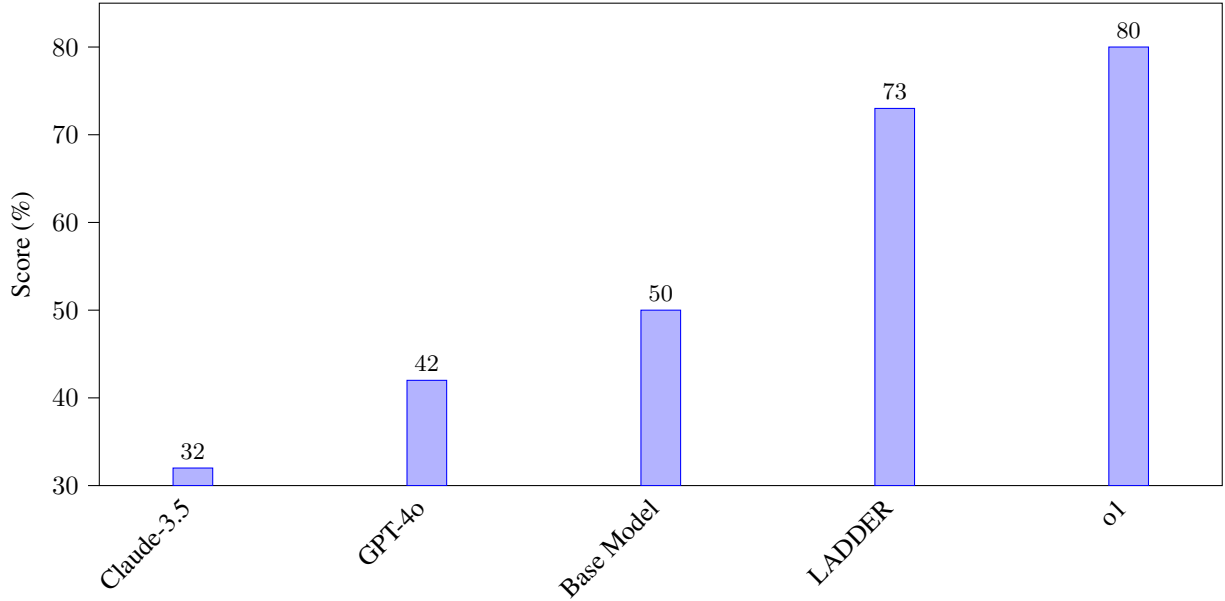

We observed a clear upward trend in performance as the number of generated variants increased, with no definitive plateau in the graph (see Figure \ref{fig:scaling-performance}). While the rate of improvement appears to slow beyond 6,000 variants, the data suggests that continued scaling may still yield further gains. This indicates that generating more variants could enhance performance, though with diminishing returns relative to earlier stages of training. Rather than identifying a strict optimal corpus size, our findings highlight a trade-off between computational cost and incremental improvements, suggesting that additional scaling may remain beneficial depending on resource availability.

The stark improvement from base model pass@10 (2\%) to LADDER (82\%) demonstrates genuine acquisition of mathematical abilities rather than merely improved output sampling. The consistent failure of RL without variants under identical hyperparameters confirms that success stems from the carefully constructed difficulty gradient rather than the RL algorithm itself, suggesting potential applications to other complex reasoning tasks where direct training proves ineffective.

\subsection{MIT integration Bee (LADDER)}

Applying our methodology to the MIT Integration Bee led to a significant improvement in model performance. LADDER improved the accuracy of the Deepseek-R1 Qwen 2.5 7B base model from a baseline of 50\% to 73\% on the 2025 qualifying examination, significantly outperforming other existing models, including GPT-4o (42\%). See Figure \ref{fig:mit-perf}.

For comparison, most human participants score between 3-6 points out of 20 (15-30\%), with scores above 7-12 (35-73\%) considered strong performances. With a 73\% accuracy, our model meets the qualification threshold of 14/20. However, while LADDER achieved substantial gains, it still did not reach the performance of o1 (80\%).

The consistent performance gap between our trained model and base models highlights the effectiveness of our variant-based recursive training approach in improving mathematical reasoning. Despite the relatively modest parameter count of 7B, our base model was able to significantly outperform larger models that did not undergo targeted recursive training. This suggests that structured self-improvement methodologies can lead to substantial gains in problem-solving ability without requiring massive increases in model size.

\subsection{MIT Integration Bee (LADDER + TTRL)}

\begin{figure}[ht]
\centering
\begin{tikzpicture}
\begin{axis}[
    akira_style,
    ybar,
    bar width=15pt,
    width=13cm,
    height=6cm,
    symbolic x coords={ LADDER pass@1, LADDER pass@100, LADDER + TTRL},
    xtick=data,
    xticklabel style={rotate=0,anchor=north},
    enlarge x limits=0.3,
    ylabel={Score (\%)},
    ymin=60,
    ymax=100,
    nodes near coords,
    nodes near coords align={above},
    every node near coord/.append style={font=\small,text=black},
    ]
\addplot coordinates {
    (LADDER pass@1, 73)
    (LADDER pass@100,80)
    (LADDER + TTRL, 90)
};

\draw[dash pattern=on 3pt off 1pt] (rel axis cs:0,0.5) -- (rel axis cs:1,0.5) 
    node [pos=0.33, anchor=south, font=\small] {o1 pass@1};

\end{axis}
\end{tikzpicture}
\caption{TTRL score improvement on 2025 MIT Integration Bee.} 
\label{fig:ttrl-perf}
\end{figure}
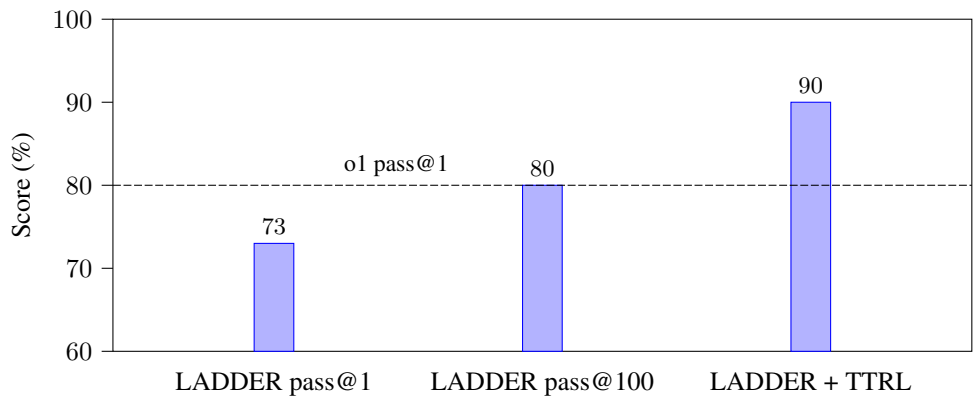

After applying LADDER to the 7B base model, we further employed Test-Time Reinforcement Learning (TTRL) for 100 steps on questions that LADDER failed to answer correctly. Of the remaining incorrect responses, TTRL successfully solved 3-4 additional problems, increasing performance from 73\% to 90\% (Figure \ref{fig:ttrl-perf}). This improvement places our model's performance well above the typical qualification threshold of 73\% for the MIT Integration Bee, setting a new state-of-the-art for mid-sized LLMs on this benchmark.

In Figure \ref{fig:ttrl-perf}, we compare LADDER pass@100 with TTRL, demonstrating that TTRL is a more effective test-time scaling method than naively attempting a test question many times. TTRL's application of recursive problem decomposition on the test questions themselves allows the initial LADDER model to correctly solve some of the hardest questions on the MIT Integration Bee qualifying examination. TTRL enables our relatively small 7B LADDER model to outperform OpenAI's o1 model at pass@1.

The number of RL steps required to solve each test question previously incorrectly answered varied, ranging from as few as 3 steps to as many as 30. This variation likely reflects the complexity of each integral, with more complex problems requiring a greater number of solved variants to develop a useful learning trajectory. The questions that remained unsolved even after LADDER + TTRL are particularly complex for undergraduate students and are included in the appendix.

To assess the necessity of LADDER in TTRL's effectiveness, we also applied TTRL directly to the base model (pre-LADDER). After 100 steps, the base model remained unable to correctly answer any of the questions that LADDER + TTRL successfully solved. Performance gains from TTRL only occurred when applied to the LADDER-trained model, suggesting that the RL protocol in LADDER provides essential exposure to the distribution of integrals, enabling TTRL to later learn further. Future research could explore whether TTRL alone, with sufficiently high compute constraints, could achieve comparable results.

We note that while our LADDER+TTRL model surpasses o1’s score on the MIT Integration Bee, this comparison should be taken as a general baseline rather than a strict head-to-head evaluation. Unlike our approach, o1 does not have access to a numerical checker, meaning it operates under different constraints. Our results highlight the effectiveness of self-improvement via recursive problem decomposition and reinforcement learning rather than suggesting a direct superiority over o1’s approach.

\section{Discussion}
Our results demonstrate a potentially transformative approach to AI self-improvement through recursive problem decomposition and solution verification. The key insight is that models can direct their own learning trajectory by generating and solving progressively more challenging problems, requiring only a formal verifier rather than human guidance or curated datasets. This represents a significant shift from traditional supervised learning or RLHF approaches, suggesting a path toward more autonomous AI systems capable of extending their own capabilities.

The effectiveness of this self-directed learning is evident in our empirical results: improvements from 1\% to 82\% on undergraduate integration problems and from 50\% to 73\% on the MIT Integration Bee, achieved with LADDER, without human feedback or model scaling. With additional variant-based learning at test-time (TTRL), we achieve a state-of-the-art score of 90\%. The 7B parameter model's ability to outperform much larger architectures suggests that improvements in AI capabilities may come not just from scaling compute or parameters, but from enabling models to structure their own learning progression.

This methodology's recursive capability makes it particularly powerful. When faced with problems beyond their current abilities, models can autonomously decompose them into simpler variants, creating their own curriculum that adapts to their evolving capabilities. The implications extend beyond mathematical reasoning - suggesting a general framework for AI systems that can bootstrap more sophisticated capabilities from simpler ones, guided only by formal verification mechanisms rather than human oversight.

\begin{figure}[t]
\centering 
\includegraphics[width=0.9\textwidth]{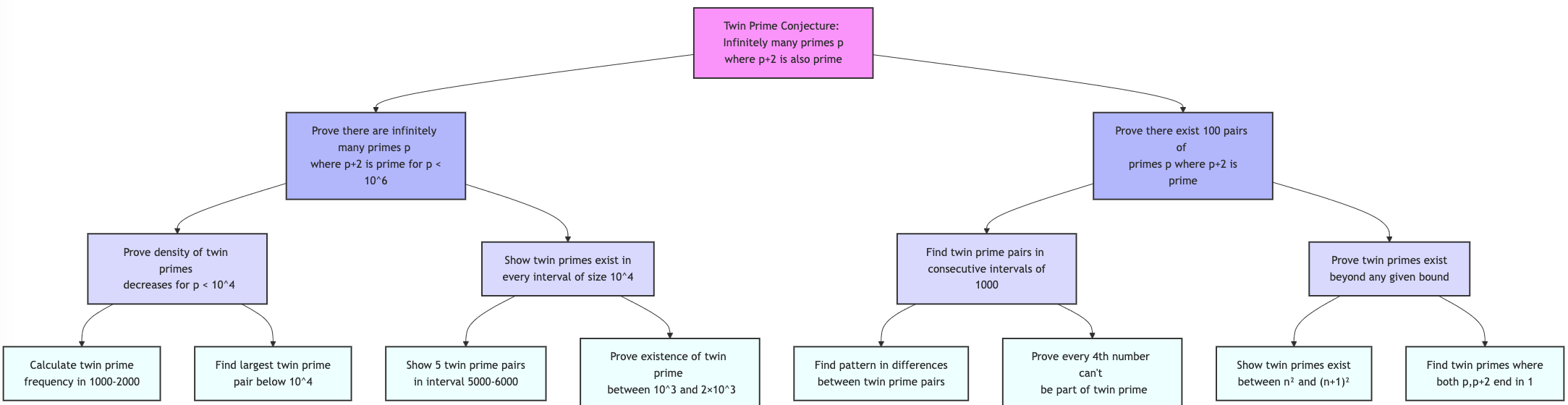}
\caption{Example of variant generation for twin prime conjecture problem. }
\label{fig:proof-variants}
\end{figure}
\subsection{A Form of Test-Time Compute Scaling}
Test-time compute scaling represents a promising new frontier in enhancing model capabilities beyond traditional approaches like architectural scaling or increased parameter counts. LADDER and TTRL introduce a fundamentally different paradigm: rather than relying on larger models or more extensive pre-training, we can achieve substantial performance improvements by enabling models to strategically practice and learn during inference. This approach is particularly powerful because it allows models to develop problem-specific expertise dynamically, adapting their capabilities to the exact challenges they encounter.

The dominant approach to test-time compute scaling today concentrates on increasing output token length, allowing models to engage in more extensive step-by-step reasoning. However, this approach faces fundamental challenges with extremely difficult problems that require extensive exploration. The sequential nature of token generation creates memory constraints and potential bottlenecks, as each token must be generated and processed in sequence while maintaining the entire chain of reasoning in context.

TTRL offers a fundamentally different approach that mirrors human learning strategies. Just as humans often need to study and practice similar problems before tackling a particularly challenging question, TTRL enables models to systematically explore and learn from related problems before attempting the target task. This approach is inherently more parallelizable than sequential token generation - variant problems can be generated and solved independently across multiple compute units, with their insights aggregated to improve performance on the original problem. While our current implementation explored variants sequentially, the framework naturally extends to distributed training approaches, potentially offering significant speedups through parallel exploration of the problem space.

This parallelizability presents a key advantage over traditional test-time scaling approaches. Where increasing output length faces fundamental sequential constraints, TTRL's variant generation and learning process can be distributed across multiple GPUs or machines. Each compute unit can independently explore different regions of the problem space, generating and solving variants while sharing insights through model updates. This mirrors distributed training approaches in the pre-training phase, but applied dynamically at test-time to specific challenging problems.

\subsection{Extension to Other Verifiable Domains}

While we demonstrated our approach using numerical integration, the underlying principle can extend to a broad range of formal reasoning tasks through appropriate verification tools. It is likely that any domain where question variants can be generated and which has a verifier-generator gap can leverage our approach. These domains share the critical property we identified in integration: a clear generator-verifier gap where solution verification is more straightforward than solution generation. This suggests our approach of iterative variant generation and verified learning could provide a general framework for improving formal reasoning capabilities in language models.

Competitive programming is an example of another possible domain in which our method may be effective. An LLM could generate a solution to a programming problem and verify it with provided unit tests. In formal mathematics, tools like the Lean proof assistant offer another avenue for verification. Lean can rigorously verify mathematical proofs, providing a reliable signal for whether a model's reasoning is correct. 

More generally, this approach has potential applications in planning and agent-based tasks, where models can teach themselves by successfully executing simpler tasks before building up to more complex challenges. This progressive learning approach, combined with reliable verification at each step, provides a robust framework for developing advanced reasoning capabilities.

\subsection{Future Work}
A core challenge in optimizing LADDER lies in the precise calibration of variant difficulty and curriculum sequencing. While our current approach uses static parameters for variant generation, an adaptive strategy could dynamically adjust the difficulty gradient based on model performance. Rather than generating a fixed number of variants at predetermined difficulty levels, the framework could analyze solution success rates to determine optimal branching points and difficulty steps. This dynamic calibration would ensure each generated variant meaningfully contributes to the learning trajectory while avoiding redundant or inappropriately difficult examples that waste computational resources. The balance between exploration of novel problem patterns and exploitation of known solution strategies remains to be optimized.

\section{Conclusion}
We proposed LADDER, a framework for improving language models' problem-solving capabilities through recursive problem decomposition and reinforcement learning with verifiable rewards, as well as its extension to test-time, TTRL. By enabling models to generate and learn from progressively simpler variants of complex problems, LADDER demonstrates how strategic self-directed learning can achieve significant improvements without relying on architectural scaling or human supervision. The dramatic improvements in mathematical reasoning capabilities - from 1\% to 82\% on undergraduate integration problems and achieving 90\% accuracy on the MIT Integration Bee through TTRL - demonstrate the effectiveness of this approach.

Beyond mathematical integration, our work builds upon three insights in general AI development: (1) the power of recursive problem decomposition for tackling complex tasks, (2) the importance of verifiable rewards for guiding self-improvement, and (3) the potential of scaling compute through strategic practice at test-time. The success of both LADDER during training and TTRL at inference time reinforces that focusing on how models interact with their environment is just as important as architectural innovations for advancing AI capabilities.

The principles demonstrated in this paper could likely extend to any domain with clear verification mechanisms, from program synthesis to formal theorem proving. By providing a framework for models to bootstrap their own capabilities through carefully constructed learning trajectories, we move closer to AI systems that can systematically extend their abilities into increasingly complex domains.

\bibliographystyle{plain}
\bibliography{references}

\section*{Appendix: MIT Integration Bee Failed Questions}

After applying LADDER and TTRL, there were two questions on the 2025 MIT Integration Bee qualifying exam which continued to be answered incorrectly. The questions incorrectly answered are among the most complex questions on the exam, and for an undergraduate student are very difficult to solve.

\begin{align*}
\text{Question 12:}\quad
& \int \sqrt[3]{x \cdot \sqrt[4]{x \cdot \sqrt[5]{x \cdot \sqrt[6]{x \cdots}}}}\,dx \\[6pt]
\text{Question 13:  }\quad
& \int \frac{e^{2x}(x^2 + x)}{(xe^x) ^4 + 1},dx  \\[6pt]
\end{align*}

\end{document}